\documentclass[10pt,twocolumn,letterpaper]{article}

\usepackage{cvpr}              

\usepackage{graphicx}
\usepackage{amsmath}
\usepackage{amssymb}
\usepackage{booktabs}
\usepackage{bbm}
\usepackage{algorithm}%
\usepackage{algorithmic}

\usepackage[pagebackref,breaklinks,colorlinks]{hyperref}
\usepackage{xcolor}

\usepackage[accsupp]{axessibility}

\def\cvprPaperID{50} 
\def\confName{CVPR}
\def\confYear{2022}

\begin{document}

\title{CNLL: A Semi-supervised Approach For Continual Noisy Label Learning }

\author{Nazmul Karim, Umar Khalid, Ashkan Esmaeili,  and Nazanin Rahnavard \\
Department of Electrical and Computer Engineering\\
University of Central Florida, USA\\
{\tt\small \{nazmul.karim18, umarkhalid, ashkan.esmaeili\}  @knights.ucf.edu} \\
{\tt\small nazanin.rahnavard@ucf.edu
}
 }
\maketitle

\begin{abstract}
The task of continual learning requires careful design of algorithms that can tackle catastrophic forgetting. However, the noisy label, which is inevitable in a real-world scenario, seems to exacerbate the situation. While very few studies have addressed the issue of continual learning under noisy labels, long training time and complicated training schemes limit their applications in most cases. In contrast, we propose a simple purification technique to effectively cleanse the online data stream that is both cost-effective and more accurate. After purification, we perform fine-tuning in a semi-supervised fashion that ensures the participation of all available samples. Training in this fashion helps us learn a better representation that results in state-of-the-art (SOTA) performance. Through extensive experimentation on 3 benchmark datasets, MNIST, CIFAR10 and CIFAR100
, we show the effectiveness of our proposed approach. We achieve a 24.8\% performance gain for CIFAR10 with 20\% noise over previous SOTA methods. Our code is publicly available.\footnote {\url{https://github.com/nazmul-karim170/CNLL}}  
\end{abstract}

\section{Introduction}
\label{sec:intro}
Deep learning models exhibit impressive performance on numerous tasks in computer vision, machine intelligence, and natural language processing \cite{9596233, refId0, khalid2022rf,khalid2022rodd}. However, deep neural networks (DNNs) struggle to \emph{continually} learn \emph{new tasks}, where the model is desired to learn sequential tasks without forgetting their previous knowledge~\cite{mittal2021essentials,peng2021defense,smith2021always}. Although several studies have addressed the issue of continual learning, the problem of \emph{continual learning} and \emph{noisy label} classification in one framework is relatively less explored. As continual learning and noisy label tasks are inevitable in real world scenarios; therefore, it is highly probable that they both emerge concurrently \cite{kim2021continual}. Hence, this study explores the development of a tangible and viable deep learning approach that can overcome both catastrophic forgetting and noisy label data challenges.

We leverage the replay-based approach to handle the continual learning tasks. However, replaying a noisy buffer degrades the performance further~\cite{kim2021continual} due to flawed mapping of the previously learned knowledge. Furthermore, existing noisy labels strategies in the literature exhibit performance limitations in the online task-free setting~\cite{kim2021continual,aljundi2019online,prabhu2020gdumb,kim2020imbalanced}. In their original framework, these methods operate on the assumption that entire dataset is provided to eliminate the noise and are thus adversely affected by a small amount of buffered data. To counter these limitations, SPR~\cite{kim2021continual} argue that if a pure replay buffer is maintained, the significant performance gains can be achieved. While SPR~\cite{kim2021continual} employs self-supervised learning in order to create a purified buffer, this type of learning demands a long training time with high computations which limits its application in practical scenario. We aim to alleviate this issue through a masking-based purification technique that is suitable for online continual learning. Furthermore, \cite{kim2021continual} only relies on clean samples and discards the potential noisy samples. While having a purified buffer offers a  better learning scenario, one can still utilize the noisy samples for unsupervised feature learning.  
\\
\begin{figure*}[t]
\centering
\includegraphics[width=1\linewidth]{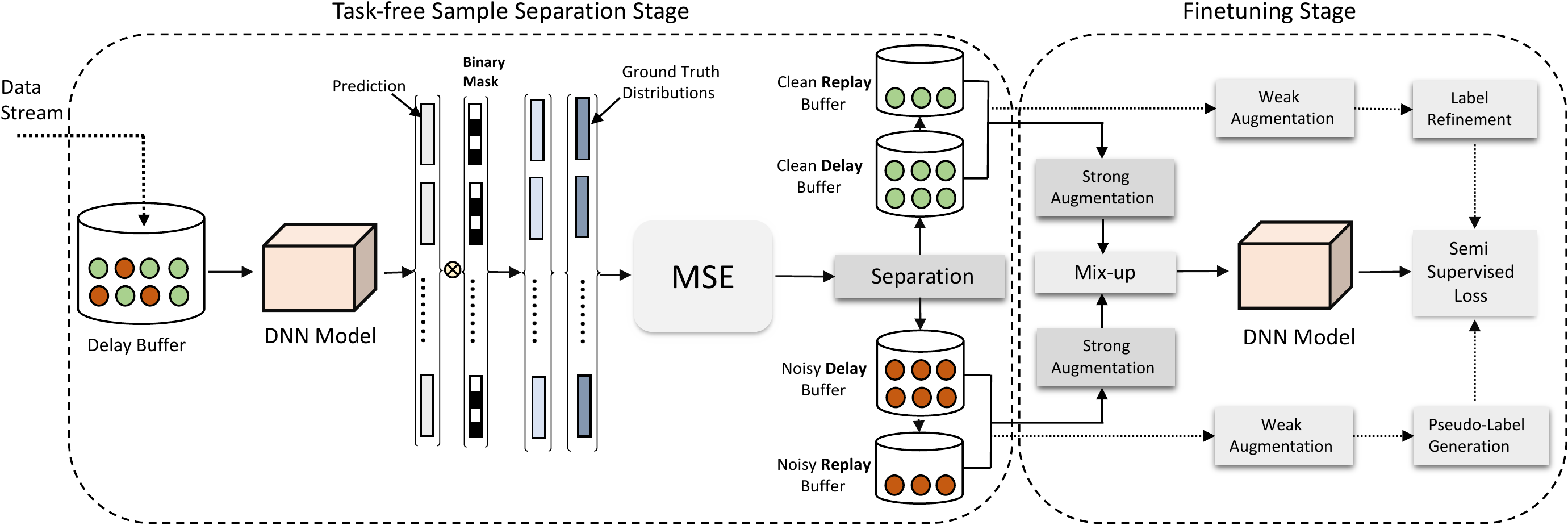}
\caption{Our proposed CNLL framework. As the data streaming continues, we warm up the model using the samples from delay buffer $\mathcal{D}$. After the warm-up, we use the mean square error ($MSE$) loss to measure the distance between predictions and ground-truth distributions. The loss distribution helps us creating the clean and noisy \emph{delay} buffers ($\mathcal{C}$ and $\mathcal{N}$). We update clean and noisy \emph{replay} buffers ($\mathcal{B_C}$ and $\mathcal{B_N}$) with samples from delay buffers. A few highly confident samples (low loss values) from $\mathcal{C}$ are stored in $\mathcal{B_C}$ while samples with low confidence are stored in $\mathcal{B_N}$. Using these buffers ($\mathcal{C}$, $\mathcal{N}$, $\mathcal{B_C}$ and $\mathcal{B_N}$), we perform an SSL-based fine-tuning that results in SOTA performance across 3 different datasets with various noise rates and types.}
\label{fig:sep}
\end{figure*}
\indent To this end, we propose a novel sample separation mechanism suitable for online task-free continual learning. We aim to circumvent the limitation of ~\cite{kim2021continual} by conventional supervised learning and a masking-based distance metric to separate the incoming data stream into two different buffers: clean and noisy buffers. Later, we employ clean buffer for supervised fine-tuning and noisy buffer for unsupervised feature learning in a semi-supervised learning (SSL) fashion. Our carefully designed separation technique with SSL-based fine-tuning achieves state-of-the-art performance. We verify our claims through extensive experimental evaluation.  

In summary, the contributions of this study can be listed as follows- 
\begin{itemize}
    \item We propose a novel purifying technique to tackle the noisy label problem in the realm of continual learning. We employ masking-based loss metric to accurately filter out clean samples from the noisy data stream.
    \item Instead of discarding the noisy samples, we propose to utilize them for unsupervised feature learning which improves the overall performance significantly. 
    \item Our proposed technique Continual Noisy Label Leaning (CNLL) outperforms current SOTA on three synthetic noise benchmarks of MNIST~\cite{lecun1998gradient}, CIFAR-10~\cite{krizhevsky2009learning}, and CIFAR-100~\cite{krizhevsky2009learning}.
\end{itemize}

\indent The  rest  of  the  paper  is  ordered as  follows. Related study is discussed in Section \ref{sec:related}. Section \ref{sec:back} highlights the background of the study, and Section \ref{sec:approach} presents the approach of the proposed noisy continual learning framework. Experimental results are illustrated in Section \ref{sec:experiments}. We discuss the scope of future work in Section~\ref{sec:future_work}. Finally, the paper is concluded in Section~\ref{sec:conclusion}.


\section{Related Work}
\label{sec:related}
The fundamental challenge of continual learning is to address the \emph{catastrophic forgetting}, where the model is trained on multiple tasks sequentially \cite{aljundi2018selfless,aljundi2019gradient,prabhu2020gdumb, caccia2020online}. In recent studies, the problem of \emph{continual learning} has been addressed mainly with four representative approaches \cite{kim2021continual}. These include regularization\cite{DBLP:journals/corr/KirkpatrickPRVD16, zenke2017continual, aljundi2018selfless}, replay \cite{8793982,kim2021continual,rolnick2019experience}, distillation \cite{li2017learning} and expansion techniques \cite{Yoon2018LifelongLW,rusu2016progressive}. The regularization based technique is designed to penalize any changes in the network parameters while learning the current task to prevent the catastrophic forgetting. EWC \cite{kirkpatrick2017overcoming}
, MAS~\cite{aljundi2018memory}, SI~\cite{zenke2017continual}, NPC~\cite{paik2020overcoming} and RWalk~\cite{chaudhry2018riemannian}  are among continual learning methods in the literature that impose regularization on the network parameters. The reply memory based methods train the model using the \emph{replayed} data of the previous tasks in addition to the data for a new task. Some of the replay-based techniques use a subset of data from the old task ~\cite{wu2019large,zhao2020maintaining}, while some others generate synthetic data to replace the original data from the previous tasks~\cite{shin2017continual,hu2018overcoming,ostapenko2019learning}. The distillation-based continual learning approaches are inspired by the knowledge-distillation~\cite{hinton2015distilling,cho2019efficacy} technique, where the model trained on the previous tasks is regarded as a teacher, and the model being trained on the new task is taken as a student. Here, the distillation loss is utilized to eliminate any performance deterioration on the previous tasks. LwF~\cite{li2017learning}, LwM~\cite{dhar2019learning}, MCIL~\cite{liu2020mnemonics} and lifelong GAN~\cite{zhai2019lifelong} are some of the popular distillation-based continual learning methods. These methods do not rely on the data from the previous tasks. They either use only the training data of the current task~\cite{li2017learning} or may generate the synthetic data~\cite{zhai2019lifelong} in addition to the training samples of the current task to train the network. In the expansion-based strategy, the model architecture is dynamically altered by expansion when new tasks are encountered. The expansion is achieved either by increasing the width or depth of the model~\cite{hung2019compacting,li2019learn,yoon2017lifelong}. As such methods may introduce computational and memory cost, an alternative technique is to employ masking the weights or neurons to break the original network into multiple sub-networks~\cite{abati2020conditional,serra2018overcoming, rajasegaran2019random}. In this scenario, each sub-network has to be trained separately for the corresponding task, and masked weights are then used, for inference. \\ 
\indent In addition, learning from data with noisy labels is another challenging task that deep learning models often encounter during training\cite{reed2014training,song2022learning,natarajan2013learning,li2017learning}. Recent researches try to overcome such training scenarios by proposing approaches such as label cleaning \cite{9190652,kim2019nlnl} and loss regularization\cite{zhang2018generalized,wang2019symmetric,hendrycks2018using} etc. Zhang et al.~\cite{zhang2021understanding} explain that deep neural networks can be efficiently trained with any ratio of noisy labels, but may generalize poorly on test samples. The established regularization methods such as weight decay, data augmentation and batch normalization have failed to overcome this overfitting issue~\cite{song2022learning}. Numerous methods, such as k-Nearest Neighbor, Anomaly Detection, and Outlier Detection have been recently studied for label cleaning that eliminate the false-labeled samples from the noisy training dataset~\cite{gamberger2000noise , delany2012profiling,thongkam2008support}. However, these methods tend to remove the clean samples from the training data and henceforth, deteriorate the overall performance of the model. Loss regularization is another technique to address the noisy label challenge that focus on designing the noise correction loss to effectively optimize the objective function only on the clean samples. Noise transition matrix has been proposed in~\cite{patrini2017making} for loss correction.~\cite{li2019learning} introduces meta-loss to optimally find parameters which are robust to noise. Bootstrapping loss has been introduced in~\cite{arazo2019unsupervised}, while~\cite{harutyunyan2020improving} proposes an information theory-based loss for loss regularization. Early learning regularization~\cite{liu2020early}, JoCoR~\cite{wei2020combating}, Jo-SRC~\cite{yao2021jo} and sparse regularization~\cite{zhou2021learning} are some of the most recent regularization approaches to combat noisy label issue. However, these methods require a lot of samples to achieve a satisfactory performance, while this may not be the case in online continual learning scenario. 

\section{Background}
\label{sec:back}
In this work, we focus on the online task-free continual learning problem. In the online case, one does not have the knowledge of a particular task's start or end times. At any time $t$, the system receives the data stream $(\mathbf{x}_t, y_t)$ drawn from a current distribution $d_{k}$. Here, $k$ indicates the task number. The distribution $d_{k}$ can experience a sudden or gradual change in $d_{k+1}$. Since the system is unaware of the occurrence of the distribution shift, it becomes challenging to prevent the catastrophic forgetting phenomenon in the online continual learning settings. Our aim is to continually learn and update a DNN model $\mathbf{f}(.;\Theta)$ such that it minimizes error on the already seen and upcoming data streams. The continuous process of accumulating and updating knowledge needs careful design of the optimization function. Given a DNN model $\mathbf{f}(.;\Theta)$ with parameters $\Theta$, the objective function we set to minimize is
\begin{equation}
    \min_{\Theta} \mathcal{L}(\mathbf{f}(X_\mathcal{D};\Theta), Y_\mathcal{D}) + \mathcal{L}(\mathbf{f}(X_\mathcal{B};\Theta), Y_\mathcal{B}). 
\end{equation}
Here, $\mathcal{L}$ is a suitable loss function. Moreover, $\{X_\mathcal{D},Y_\mathcal{D}\}$ and $\{X_{\mathcal{B}},Y_{\mathcal{B}}\}$ are the data stream sets in the current buffer $\mathcal{D}$ and a replay buffer $\mathcal{B}$. The replay buffer $\mathcal{B}$ contains a small number of selected samples from already seen data streams.  
Now, consider the data stream  $(\mathbf{x}_t, y_t)$ to contain label noise, where $y_t$ may have been mislabelled. Therefore, both $\mathcal{D}$ and $\mathcal{B}$ may contain label noise that have adverse effect on the generalization performance of the model. It has been shown in \cite{kim2021continual} that the presence of noisy label deteriorates the performance of continual learning, mirroring the effect of retrograde amnesia \cite{squire1981two}. The effect of catastrophic forgetting \cite{mccloskey1989catastrophic,ratcliff1990connectionist} is much worse under the effect of noisy labels. The reason for this is assumed to be the corrupted buffer/memory that hinders the subsequent learning. Previous study \cite{kim2021continual} shows that having a clean buffer helps in improving the performance significantly. 

\begin{algorithm}[tb]
    \caption{Task-free Sample Separation for CNLL}
    \label{alg:train_algo}
    \begin{algorithmic}
        \STATE {\bfseries Input:} Training data $(\mathbf{x}_t, y_t),...,(\mathbf{x}_{T_{train}}, y_{T_{train}})$, Warmup Period $E$,  and network parameters $\Theta$. %
        \STATE $\mathcal{C} = \mathcal{N} = \{\}$ \textit{ // Initialize Clean and Noisy Delay Buffer}
        \STATE $\mathcal{B_C} = \mathcal{B_N} = \{\}$ \textit{ // Initialize Clean and Noisy Replay Buffer}
        \STATE $\mathcal{D} = \{\}$ \textit{// Initialize Delay buffer}
        \FOR{$t=1$ {\bfseries to} $T_{train}$}
        \IF{$\mathcal{D}$ is full}
            \STATE  $\Theta\leftarrow$ $\textsc{WarmUp}(\Theta, \mathcal{D}, E)$ \
            \FOR{$i=1$ {\bfseries to} $|\mathcal{D}|$}
                \STATE $\mathbf{\mathbf{p}}_i = \mathbf{f}(\mathbf{x}_i;\Theta)$ \
                \STATE Initialize binary mask $\mathbf{m}_i$
                \STATE $l_i = MSE(\mathbf{y}_i,\mathbf{m}_i\odot\mathbf{p}_i)$
            \ENDFOR
        \STATE Calculate $l_{threshold}$ using Eq. \ref{eq:filter_rate}
        \STATE $\mathcal{D}_{clean} \leftarrow  \{ (x_i,y_i): \forall \hspace{0.2cm} l_i < l_{threshold} \} $ \\
        \STATE $\mathcal{C} \leftarrow \mathcal{C}  \ \cup \mathcal{D}_{clean}$
        \STATE $\mathcal{D}_{noisy} \leftarrow  \mathcal{D} \setminus \mathcal{D}_{clean}$
        \STATE $\mathcal{N} \leftarrow \mathcal{N}  \ \cup \mathcal{D}_{noisy}$
        \STATE Reset $\mathcal{D}$
        \ELSE
        \STATE update $\mathcal{D}$ with $(\mathbf{x}_t, y_t)$
        \ENDIF
        \IF{$\mathcal{C}$ is full}
            \STATE Update $\mathcal{B_{C}}$ using eq. \ref{eq:clean_replay}
            \STATE Reset $\mathcal{C}$
        \ENDIF
        \IF{$\mathcal{N}$ is full}
            \STATE Update $\mathcal{B_{N}}$ using eq. \ref{eq:noisy_replay}
            \STATE Reset $\mathcal{N}$
        \ENDIF
        \IF{$\mathcal{C}$ and $\mathcal{N}$ is full}
            \STATE Fine-tune and Inference Phase for $\Theta$
        \ENDIF        
        \ENDFOR
    \end{algorithmic}
\end{algorithm}

\section{Proposed Method}
\label{sec:approach}
Figure \ref{fig:sep} shows the our proposed framework where a delay/current buffer $\mathcal{D}$ of limited size stores the incoming data stream. Our objective is to separate this buffer into a clean buffer $\mathcal{C}$ and a noisy buffer $\mathcal{N}$. The former will hold the clean samples where noisy samples will be in the later. 
In contrast to SPR~\cite{kim2021continual}, we aim to learn better representations utilizing not only pure samples but also the noisy ones. We can use unsupervised feature learning for the noisy data that eventually improves the overall performance. Furthermore, SPR~\cite{kim2021continual} employs a self-supervised learning based purifying technique that requires complicated formulation and long training time. Instead, we use a much simpler, faster and effective purifying technique that seems to outperform \cite{kim2021continual} in all benchmark datasets. 

\subsection{Sample Separation}
Whenever $\mathcal{D}$ is full, we perform a warm-up of the model $\mathbf{f}(.;\Theta)$. In general, warm-up indicates a brief period of fully supervised training on $\mathcal{D}$. In this period, we minimize standard cross-entropy (CE) loss with a very low learning rate to perform warm-up. It has been shown in \cite{arpit2017closer} that DNN learns the simple pattern first before memorizing the noisy labels over the exposure of long training. Due to this fact, clean samples tends to have low loss compared to noisy samples after the warm-up. One should be able to separate $\mathcal{D}$ into clean ($\mathcal{C}$) and noisy ($\mathcal{N}$) buffers by putting a threshold on the loss values. 

\indent For $\mathbf{x}_i$, the average prediction probabilities can be denoted as $\mathbf{p}_i = [\mathbf{p}_{i}^{1}, \mathbf{p}_{i}^{2}, ...  ...,\mathbf{p}_{i}^{c}]$ and $\mathbf{y}_i = [\mathbf{y}_{i}^{1}, \mathbf{y}_{i}^{2}, ...  ...,\mathbf{y}_{i}^{c}]$ is the given ground-truth label distribution. Here, $c$ is the number of total classes in all tasks combined. To measure the differences between $\mathbf{y}_i$ and $\mathbf{p}_i$, we use mean square error ($MSE$) loss, $l_i$, which can be defined as, 
\begin{equation}\label{eq:JSD}
\begin{split}
    l_i & = MSE(\mathbf{y}_i, \mathbf{m} \odot\mathbf{p}_i) \\ 
    & =\sum_{j=1}^c (\mathbf{y}^{j}_i-\mathbf{m}^{j}_i \odot \mathbf{p}^{j}_i))^2.
\end{split}
\end{equation}
Here, $\mathbf{m}_i \in [0,1]^c$ is the class-specific binary mask that contains 1 for classes that are currently present in the delay buffer and 0 for rest of the classes. Note that $\mathcal{D}$ may not contain samples from all classes at any given time and the number of classes present in $\mathcal{D}$, $c_{\mathcal{D}}$, may vary over time. Therefore, considering prediction probabilities ($\mathbf{p}_i$) for all possible classes may result in losses that are misguided. As $c_{\mathcal{D}} \leq c$, dynamically adjusting the binary mask according to the present classes is justified and seems to alleviate this issue. After measuring $\mathbf{l} =\{l_i: i \in (1, \dots, N) \}$, we estimate the separation threshold $l_{threshold}$ as,  
\begin{equation}\label{eq:filter_rate}
  l_{threshold} = l_{mean}, 
\end{equation}
where $l_{mean}$ is the mean of the loss distribution $\mathbf{l}$ and $N$ is the number of samples in delay buffer $\mathcal{D}$.

\begin{algorithm}[!t]
    \caption{Fine-tuning and Inference Stage of CNLL}
    \label{alg:eval_algo}
    \begin{algorithmic}
        \STATE {\bfseries Input:} Test data $(\mathbf{x}_t, y_t),...,(\mathbf{x}_{T_{test}}, y_{T_{test}})$, network parameters $\Theta$, clean and noisy delay buffers $\mathcal{C}$ and $\mathcal{N}$, clean and noisy replay buffers $\mathcal{B_C}$ and $\mathcal{B_N}$.
        \STATE $\Psi\leftarrow\Theta$ \hspace{3.5cm} \textit{//Copy parameters}
        \STATE Get $\mathcal{S}$ and $\mathcal{U}$ sets using eq. \ref{eq:lab_sets}
        \STATE $\Psi\leftarrow$ FixMatch($\mathcal{S}$, $\mathcal{U}$,  $\Psi$) \hspace{0.3cm} \textit{//SSL Fine-tuning Phase}
        \FOR{$t=1$ {\bfseries to} $T_{test}$}
        \STATE Classification of $x_t$ using $\Psi$ \hspace{0.2cm} \textit{//Inference Phase}
        \ENDFOR
    \end{algorithmic}
\end{algorithm}  
Our proposed thresholding method neither depends on any type of training hyper-parameters, nor needs to be adjusted for different noise type, rate, or even datasets. We empirically validate that taking $l_{mean}$ as the threshold gives us the best separation of clean and noisy samples. Furthermore, whenever $\mathcal{C}$ is full, we hold $N_1$ number of highly confident (with low loss values) clean samples from $C$ in the clean replay buffer $\mathcal{B_C}$. On the contrary, $N_2$ number of noisy samples with high loss values are stored in the noisy replay buffer $\mathcal{B_N}$. Considering $\mathbf{l}_{\mathcal{C}}$ and $\mathbf{l}_{\mathcal{N}}$ are loss vectors containing loss values for samples in $\mathcal{C}$ and  $\mathcal{N}$. The update rule for the clean replay buffer $\mathcal{B_C}$ can be expressed as, 
\begin{equation}\label{eq:clean_replay}
\begin{split}
    \mathbf{l}_{\mathcal{C}}^{(low)} \leftarrow \text{Lowest} \hspace{0.1cm} N_1 \hspace{0.1cm} \text{values in} \hspace{0.1cm} \mathbf{l}_{\mathcal{C}} \\
    \mathcal{B_C} \leftarrow  \{(x_i,y_i):  \forall \hspace{0.2cm} l_i \in \mathbf{l}_{\mathcal{C}}^{(low)}\} 
\end{split}
\end{equation}
Similarly for the noisy replay buffer $\mathcal{B_N}$, we set the update rule as
\begin{equation} \label{eq:noisy_replay}
    \begin{split}
    \mathbf{l}_{\mathcal{N}}^{(high)} \leftarrow \text{Highest} \hspace{0.1cm} N_2 \hspace{0.1cm} \text{values in} \hspace{0.1cm} \mathbf{l}_{\mathcal{N}} \\
    \mathcal{B_N} \leftarrow  \{(x_i,y_i):  \forall \hspace{0.2cm} l_i \in \mathbf{l}_{\mathcal{N}}^{(high)}\} 
    \end{split}
\end{equation}

Algorithm \ref{alg:train_algo} summarizes the proposed sample separation approach.

\subsection{Fine-tuning Stage}
Figure \ref{fig:sep} also shows the fine-tuning stage after separation. Since the labels of the noisy buffers ($\mathcal{N}$ and $\mathcal{B_N}$) cannot be trusted, we consider them as the unlabeled data. On the other hand, clean buffers ($\mathcal{C}$ and $\mathcal{B_C}$) contain data with reliable labels. Therefore, these buffers can be considered as the source of labeled data. To this end, we can define the labelled and unlabelled sets, $\mathcal{S}$ and $\mathcal{U}$ for the SSL-based fine-tuning as
\begin{equation}\label{eq:lab_sets}
\begin{split}
    \mathcal{S} & = \mathcal{C} \cup \mathcal{B_C}\\
    \mathcal{U} & = \mathcal{N} \cup \mathcal{B_N}
\end{split}
\end{equation}
We consider SSL-based training utilizing both labeled and unlabeled data. We follow the FixMatch~\cite{sohn2020fixmatch} for SSL training. At first, we generate two weakly augmented copies of samples from $\mathcal{S}$ and $\mathcal{U}$. The model generated predictions for both these copies are $\mathbf{q}_{t}$ and $\mathbf{r}_{t}$. For labeled data, we refine their labels $\mathbf{y}_t$ using the model prediction $\mathbf{q}_{t}$ as, 
\begin{equation} \label{eq:label_refinement}
    \bar{\mathbf{y}}_t = w_t \mathbf{y}_t + (1-w_t)\mathbf{q}_{t}, 
\end{equation}
where $w_t$ is the label refinement coefficient. The pseudo-label $\mathbf{r}_{t}$ for each of the samples in $\mathcal{U}$ are produced solely through model's prediction.
\begin{table*}[t]
    \centering
    \begin{tabular}{lccc|cc|ccc|cc}
        \toprule
        &\multicolumn{5}{c}{MNIST}         &\multicolumn{5}{c}{CIFAR-10}\\
        &\multicolumn{3}{c}{symmetric}    &\multicolumn{2}{c}{asymmetric} &\multicolumn{3}{c}{symmetric} &\multicolumn{2}{c}{asymmetric}\\
        Noise rate (\%)  & 20 & 40 & 60 & 20 & 40 & 20 & 40 & 60  & 20 & 40 \\
        \midrule
        Multitask 0\% noise~\cite{caruana1997multitask} & \multicolumn{5}{c|}{98.6} &\multicolumn{5}{c}{84.7}  \\

       \specialrule{0.1pt}{1pt}{1pt}
        Finetune  & 19.3 & 19.0 & 18.7 & 21.1 & 21.1 & 18.5 & 18.1 & 17.0 & 15.3 & 12.4  \\
        \specialrule{0.1pt}{1pt}{1pt}
        EWC~\cite{DBLP:journals/corr/KirkpatrickPRVD16}  & 19.2 & 19.2 & 19.0 & 21.6 & 21.1 & 18.4 & 17.9 & 15.7 & 13.9 & 11.0  \\
        \specialrule{0.1pt}{1pt}{1pt}
        CRS~\cite{vitter1985random} & 58.6 & 41.8 & 27.2 & 72.3 & 64.2 & 19.6 & 18.5 & 16.8 & 28.9 & 25.2   \\
        CRS + L2R~\cite{ren2018learning} & 80.6 & 72.9 & 60.3 & 83.8 & 77.5 &  29.3 & 22.7 & 16.5 & 39.2 & 35.2 \\
        CRS + Pencil~\cite{yi2019probabilistic} & 67.4 & 46.0 & 23.6 & 72.4 & 66.6 & 23.0 & 19.3 & 17.5 & 36.2 & 29.7   \\
        CRS + SL~\cite{wang2019symmetric} & 69.0 & 54.0 & 30.9 & 72.4 & 64.7 & 20.0 & 18.8 & 17.5 & 32.4 & 26.4 \\
        CRS + JoCoR~\cite{wei2020combating} & 58.9 & 42.1 & 30.2 & 73.0 & 63.2 & 19.4 & 18.6 & 21.1 & 30.2 & 25.1  \\

        \specialrule{0.1pt}{1pt}{1pt}
        PRS~\cite{kim2020imbalanced} & 55.5 & 40.2 & 28.5 & 71.5 & 65.6 & 19.1 & 18.5 & 16.7 & 25.6 & 21.6  \\
        PRS + L2R~\cite{ren2018learning} & 79.4 & 67.2 & 52.8 & 82.0 & 77.8 & 30.1 & 21.9 & 16.2 & 35.9 & 32.6 \\
        PRS + Pencil~\cite{yi2019probabilistic} & 62.2 & 33.2 & 21.0 & 68.6 & 61.9 & 19.8 & 18.3 & 17.6 & 29.0 & 26.7   \\
        PRS + SL~\cite{wang2019symmetric} & 66.7 & 45.9 & 29.8 & 73.4 & 63.3 & 20.1 & 18.8 & 17.0 & 29.6 & 24.0  \\
        PRS + JoCoR~\cite{wei2020combating} & 56.0 & 38.5 & 27.2 & 72.7 & 65.5 & 19.9 & 18.6 & 16.9 & 28.4 & 21.9  \\

        \specialrule{0.1pt}{1pt}{1pt}
        MIR~\cite{aljundi2018memory}  & 57.9 & 45.6 & 30.9 & 73.1 & 65.7 & 19.6 & 18.6 & 16.4 & 26.4 & 22.1  \\
        MIR + L2R~\cite{ren2018learning} & 78.1 & 69.7 & 49.3 & 79.4 & 73.4 & 28.2 & 20.0 & 15.6 & 35.1 & 34.2   \\
        MIR + Pencil~\cite{yi2019probabilistic} & 70.7 & 34.3 & 19.8 & 79.0 & 58.6 & 22.9 & 20.4 & 17.7 & 35.0 & 30.8   \\
        MIR + SL~\cite{wang2019symmetric} & 67.3 & 55.5 & 38.5 & 74.3 & 66.5 & 20.7 & 19.0 & 16.8& 28.1 & 22.9  \\
        MIR + JoCoR~\cite{wei2020combating} & 60.5 & 45.0 & 32.8 & 72.6 & 64.2 & 19.6 & 18.4 & 17.0 & 27.6 & 23.5 \\

        \specialrule{0.1pt}{1pt}{1pt}
        GDumb~\cite{prabhu2020gdumb}  & 70.0 & 51.5 & 36.0 & 78.3 & 71.7 & 29.2 & 22.0 & 16.2 & 33.0 & 32.5  \\
        GDumb + L2R~\cite{ren2018learning} & 65.2 & 57.7 & 42.3 & 67.0 & 62.3 & 28.2 & 25.5 & 18.8 & 30.5 & 30.4 \\
        GDumb + Pencil~\cite{yi2019probabilistic} & 68.3 & 51.6 & 36.7 & 78.2 & 70.0 & 26.9 & 22.3 & 16.5 & 32.5 & 29.7 \\
        GDumb + SL~\cite{wang2019symmetric} & 66.7 & 48.6 & 27.7 & 73.4 & 68.1 & 28.1 & 21.4 & 16.3 & 32.7 & 31.8  \\
        GDumb + JoCoR~\cite{wei2020combating} & 70.1 & 56.9 & 37.4 & 77.8 & 70.8 & 26.3 & 20.9 & 15.0 & 33.1 & 32.2 \\
        \specialrule{0.7pt}{1pt}{1pt}

        SPR~\cite{kim2021continual} & 85.4 & 86.7  & 84.8  & 86.8 & 86.0    & 43.9  & 43.0  & 40.0  & 44.5  & 43.9         \\
        \specialrule{0.7pt}{1pt}{1pt}

        CNLL(ours)            & \textbf{92.8}          & \textbf{90.1}          & \textbf{88.8}          & \textbf{91.5}                  & \textbf{89.4}            & \textbf{68.7}          & \textbf{65.1}          & \textbf{52.8}          & \textbf{67.2}          & \textbf{59.3}  \\
        \bottomrule
    \end{tabular}
    \caption{Overall accuracy for MNIST and CIFAR-10 datasets. Results are generated after noisy labeled continual learning for all tasks as explained in \ref{sec:approach}. Here, the delay buffer size is 500 for both datasets. The results reported here are the average of five experiments conducted with unique random seed. Results for other methods are presented according to \cite{kim2021continual}.}
    \label{tab:exp_accuracy}
\end{table*}
\begin{table*}[htb]
    \centering
    \scalebox{1}{
    \begin{tabular}{lccc|ccc}
        \toprule
        &\multicolumn{3}{c|}{random symmetric} &\multicolumn{3}{c}{superclass symmetric} \\
        \midrule
        noise rate (\%)  & 20 & 40 & 60 & 20 & 40 & 60 \\
        \midrule
        GDumb + L2R~\cite{ren2018learning} & 15.7 & 11.3 & 9.1 & 16.3 & 12.1 & 10.9 \\
        GDumb + Pencil~\cite{yi2019probabilistic} & 16.7 & 12.5 & 4.1 & 17.5 & 11.6 & 6.8 \\
        GDumb + SL~\cite{wang2019symmetric} & 19.3 & 13.8 & 8.8 & 18.6 & 13.9 & 9.4 \\
        GDumb + JoCoR~\cite{wei2020combating}  & 16.1 & 8.9 & 6.1 & 15.0 & 9.5 & 5.9 \\
        SPR~\cite{kim2021continual} & 21.5 & 21.1 & 18.1 & 20.5 & 19.8 & 16.5 \\
        \midrule
        CNLL(ours) & \textbf{38.7} & \textbf{32.1} & \textbf{26.2} & \textbf{39.0} & \textbf{32.6} & \textbf{27.5}  \\
        \bottomrule
    \end{tabular}}
    \caption{Results reported here are obtained after noisy labeled continual learning on CIFAR-100. The results reported here are the average of five experiments conducted with unique random seed. Results for other methods are presented according to \cite{kim2021continual}.}
    \label{tab:exp_cifar100}
\end{table*}

\begin{table*}[htb]
    \centering
    \scalebox{1}{
    \begin{tabular}{lccc|cc|ccc|cc}
        \toprule
        &\multicolumn{5}{c}{MNIST}  &\multicolumn{5}{c}{CIFAR-10} \\
        \midrule
        &\multicolumn{3}{c|}{symmetric} &\multicolumn{2}{c|}{asymmetric} &\multicolumn{3}{c|}{symmetric} &\multicolumn{2}{c}{asymmetric} \\
        noise rate (\%)  & 20 & 40 & 60 & 20 & 40 & 20 & 40 & 60  & 20 & 40 \\
        \midrule
        AUM~\cite{pleiss2020identifying} & 7.0 & 16.0 & 11.7 & 30.0 & 29.5 & 36.0 & 24.0 & 11.7 & 46.0 & 30.0 \\
        %
        INCV~\cite{chen2019understanding} & 23.0 & 22.5 & 14.3 & 37.0 & 31.5 & 22.0 & 18.5 & 9.3 & 37.0 & 30.0  \\
        %
        SPR~\cite{kim2021continual}  & 96.0 & 96.5 & 93.0 & 100 & 96.5 & 75.5 & 70.5 & 54.3 & 69.0 & 60.0 \\
        \midrule
        CNLL(ours)  & \textbf{98.1} & \textbf{98.0} & \textbf{96.8} & \textbf{100} & \textbf{98.6} & \textbf{82.6} & \textbf{80.1} & \textbf{66.0} & \textbf{81.3} & \textbf{73.5} \\        
        \bottomrule
    \end{tabular}}
    \vskip 0.1in
    \caption{Filtered noisy label percentage in the clean buffer. CNLL is significantly better at separating the clean samples from the noisy samples.}
    \label{tab:exp_filtering}
\end{table*}

\begin{figure*}[htb]
\centering
\begin{subfigure}{0.48\linewidth}
    \includegraphics[width=0.95\linewidth]{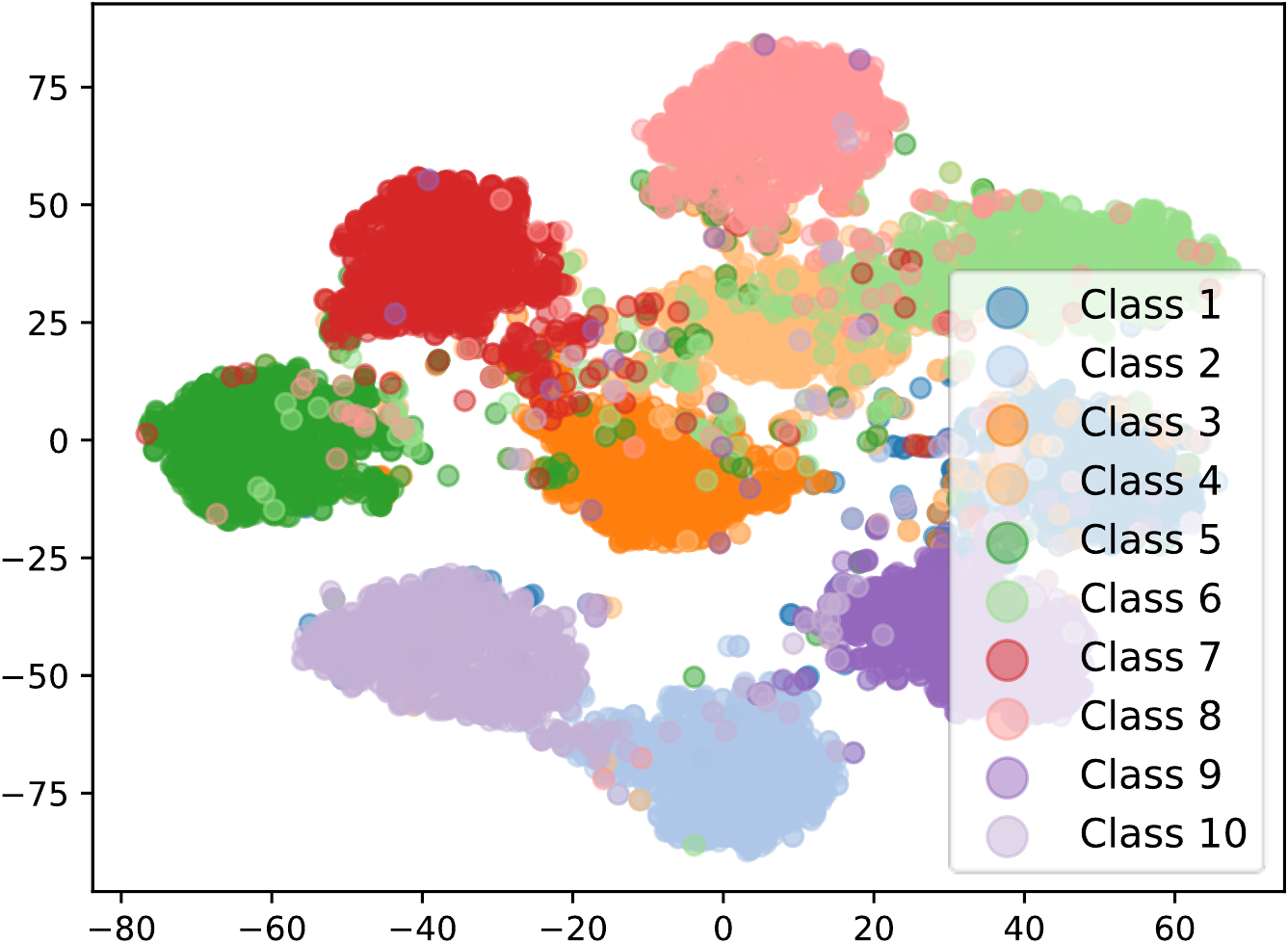}
    \caption{Embeddings of Test Samples at 0\% Noise Rate}
\end{subfigure}
\begin{subfigure}{0.48\linewidth}
    \includegraphics[width=0.95\linewidth]{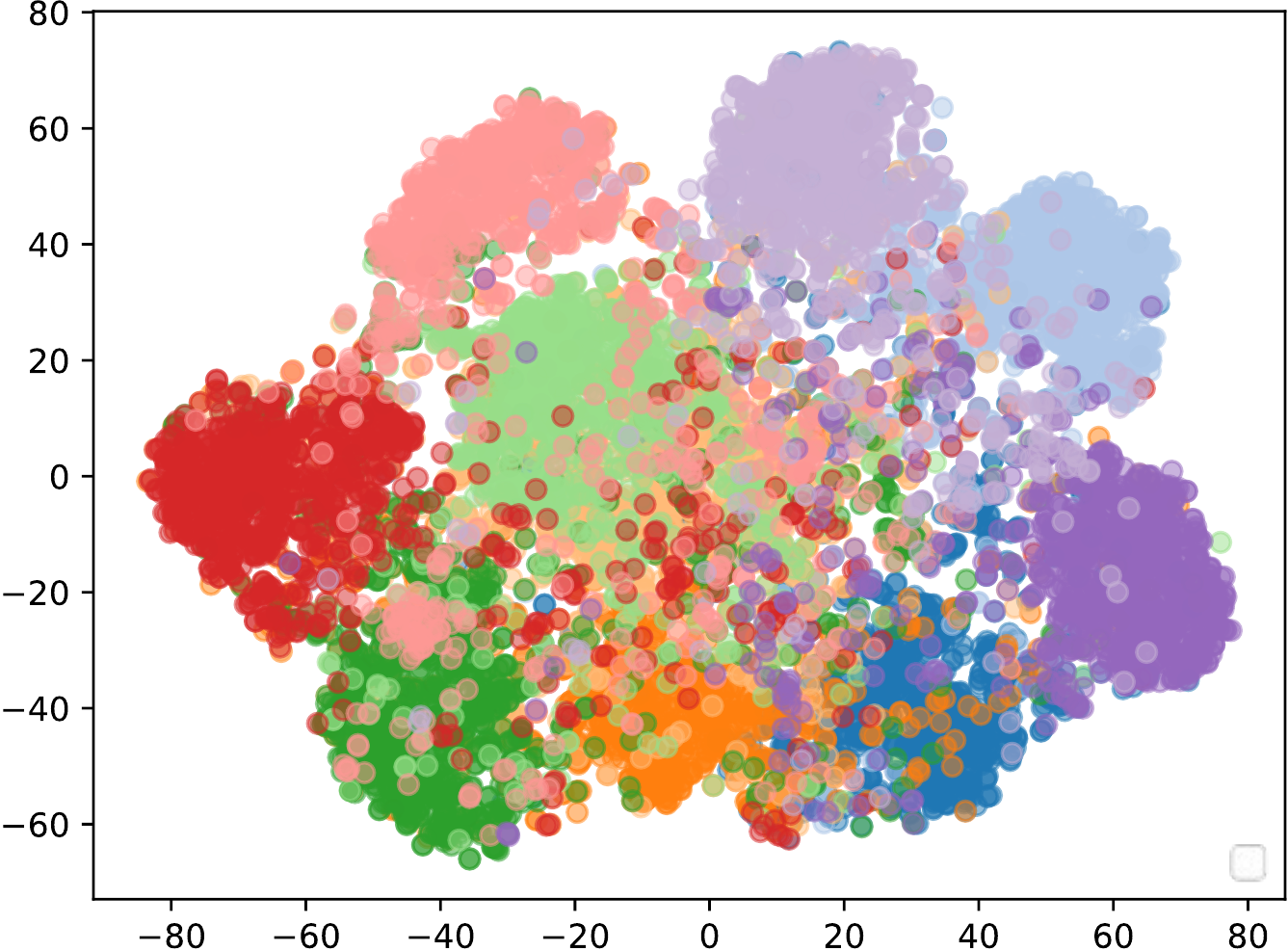}
    \caption{Embeddings of Test Samples at 20\% Noise Rate}
\end{subfigure}
\caption{t-SNE visualizations of the output features extracted for CIFAR-10 test dataset. (a) Model is trained using 0\% of noisy labels. (b) Model is trained using 20\% of noisy labels. Even under noisy labels, CNLL can separate the test samples in a satisfactory manner.}
\label{fig:tsne}
\end{figure*}
To this end, we apply label-preserving augmentation technique, Mixup~\cite{zhang2017mixup}, to strongly augmented copies of $\mathcal{S}$ and $\mathcal{U}$. Finally, the semi-supervised loss function we minimize is 
\begin{equation} \label{eq:combined_loss}
    \mathcal{L}=\mathcal{L}_\mathcal{X}+ \lambda_\mathcal{U} \mathcal{L}_\mathcal{U} + \lambda_{r} \mathcal{L}_\mathrm{reg}.    
\end{equation}
Here, $\mathcal{L}_\mathcal{X}$ and $\mathcal{L}_\mathcal{U}$ are the loss functions for labelled and unlabelled data. Moreover, $\lambda_\mathcal{U}$ and $\lambda_r$ are unsupervised loss coefficient and regularization coefficient, respectively. We employ the regularization loss $\mathcal{L}_\mathrm{reg}$  is to prevent single-class assignment of all samples. We define it based on a prior uniform distribution $(\pi_c = 1/c)$ to regularize the network's output across all samples in the mini-batch similar to Tanaka et al. \cite{tanaka2018joint} ,
\begin{equation}
    \mathcal{L}_{reg} = \sum_c {\pi_c log\big( \frac{\pi_c}{\frac{1}{|\mathcal{S}+\mathcal{U}|} \sum_{\mathbf{x} \in |\mathcal{S}+\mathcal{U}|} \mathbf{f}(\mathbf{x}; \Theta))} \big)}
\end{equation}
We describe the fine-tuning and testing stages in Algorithm \ref{alg:eval_algo}.

\section{Experiments}
\label{sec:experiments}
In this section, we draw comparison between CNLL and other SOTA models in the settings proposed in SPR~\cite{kim2021continual}. The proposed method is evaluated on three benchmarks datasets MNIST~\cite{lecun1998gradient}, CIFAR-10~\cite{krizhevsky2009learning}, and CIFAR-100~\cite{krizhevsky2009learning}.

\subsection{Experimental Design}
We design our experiments based on the recent works for robust evaluation in continual learning~\cite{kim2021continual,aljundi2018selfless,van2019three}. We consider five tasks on CIFAR-10~\cite{krizhevsky2009learning} and MNIST~\cite{lecun1998gradient}, and 2  random classes for each task. On the other hand, 20 tasks are considered on CIFAR-100, where each task has 5 classes chosen in 2 different ways: a) according to super-classes, and b) randomly. 
For CIFAR10 and MNIST, we use two noise models to create the synthetic noise dataset. First, we employ \emph{symmetric label noise} where some portion of samples from a particular class are uniformly distributed over other classes. Five tasks are then formed by picking class pairs randomly without replacement. Secondly, the \emph{asymmetric label noise} is introduced by allocating other similar class labels \cite{li2019learning}. For symmetric noise, we consider 3 different noise rates of 20\%, 40\%, and 60\%. We consider only 20\%, 40\% rates for aysmmetric case. The warmup period lasts for 30 epochs whereas we perform fine-tuning for 60 epochs. During warmup, we employ a learning rate of 0.001 with a batch size of 64. As there is noise in the data stream, higher learning rate may result in memorization of the noisy samples. Which in turn will create faulty separation of samples leading to poor generalization performance. Whereas, we amplify the learning rate to 0.1 for the fine-tuning stage. As for buffers size, we make the $\mathcal{B_C}$ and $\mathcal{B_N}$ variable length buffers. The values of $T_{train}$ and $T_{test}$ are also variables and depends on the duration of the training and inference period. For CIFAR10, we consider a pair of values of $50,000$ and $10,000$ for $T_{train}$ and $T_{test}$. Details of other hyper-parameters can be found in Table \ref{tab:hyper}.

\begin{table}[htb]
    \centering
    \begin{tabular}{c|c} \toprule
    Hyper-Parameter & Value \\ 
    \midrule
    Size of $\mathcal{D}$ & 500 \\
    Size of $\mathcal{C}$ & 500 \\
    Size of $\mathcal{N}$ & 1000 \\
    $N_1$ & 25 \\
    $N_2$ & 50 \\
    SGD Momentum & 0.9 \\
    Weight Decay & $5e^{-4}$ \\
    \bottomrule
    \end{tabular}
    \caption{Hyper-parameters used for training.}
    \label{tab:hyper}
\end{table}

As for model architectures, MLP~\cite{tang2015extreme} with two hidden layers is used for all MNIST experiments, and ResNet-18~\cite{he2016identity} architecture is used for CIFAR-10,  CIFAR-100 experiments.

\subsection{Baseline Methods}
As mentioned earlier, this study explores continual learning scenario with noisy labeled data. Therefore, the baseline is designed by incorporating SOTA methods proposed for noisy labels learning and continual learning. For continual learning, we choose CRS~\cite{riemer2018learning}, MIR~\cite{aljundi2019online}, PRS~\cite{kim2020imbalanced} and Gdumb~\cite{prabhu2020gdumb}. Six approaches are selected from the literature for the noisy label learning that are SL~\cite{wang2019symmetric},JoCoR~\cite{wei2020combating}, L2R~\cite{ren2018learning}, Pencil~\cite{yi2019probabilistic}, AUM~\cite{pleiss2020identifying}, and INCV~\cite{chen2019understanding}. Furthermore, CNLL is evaluated against SPR \cite{kim2021continual} which in our knowledge is the only study that has addressed continual learning and noisy labels concurrently. 

The hyperparameters for the baselines are as follows.
\begin{itemize}
    \item Multitask~\cite{caruana1997multitask}: We perform i.i.d offline training for 50
epochs with uniformly sampled mini-batches.
    \item Finetune: We run online training through the sequence
of tasks.
    \item GDumb~\cite{prabhu2020gdumb} : As an advantage to GDumb, we allow
CutMix with $p = 0.5$ and $\alpha = 1.0$. We use the
SGDR schedule with $T_{0} = 1$ and $T_{mult} = 2$.
Since access to a validation data in task-free continual learning is not natural, the number of epochs is set
to 100 for MNIST and CIFAR-10 and 500 for WebVision.
    \item PRS~\cite{kim2020imbalanced}: We set $\rho = 0$.
    \item L2R~\cite{ren2018learning}: We use meta update with $\alpha = 1$, and set
the number of clean data per class as 100 and the clean
update batch size as 100.
    \item Pencil~\cite{yi2019probabilistic}: We use $\alpha = 0.4, \beta = 0.1$, stage1 = 70,
stage2 = 200, $\lambda = 600$.
    \item SL~\cite{wang2019symmetric}: We use $\alpha = 1.0, \beta = 1.0$.
    \item JoCoR~\cite{wei2020combating}: We set $\lambda = 0.1$.
    \item AUM~\cite{pleiss2020identifying}: We set the learning rate to 0.1, momentum
to 0.9, weight decay to 0.0001 with a batch size of 64
for 150 epochs. We apply random crop and random
horizontal flip for input augmentation.
    \item INCV~\cite{chen2019understanding}: We set the learning rate to 0.001, weight
decay to 0.0001, a batch size 128 with 4 iterations for
200 epochs. We apply random crop and random horizontal flip for input augmentation.
    \item SPR~\cite{kim2021continual}: Self-supervised batch size is 300 for MNIST and 500 for CIFAR-10. Among other parameters, we set $\beta_1 = 0.9, \beta_2 = 0.999, \epsilon = 0.0002, E_{max} = 5$. 
\end{itemize}

\subsection{Results}
For CIFAR-10 and MNIST, the performance of CNLL has been compared with baselines in Table \ref{tab:exp_accuracy}. CNLL outperforms other methods for both symmetric and asymmetric noise types. We set the upper bound using Multitask~\cite{caruana1997multitask} which is trained under an optimal setting with perfectly clean data (i.e., the 0\% noise rate) and offline training. For 20$\%$ symmetric noise , CNLL achieves an performanc gain of 7.4\% for MNIST and 24.8\% for CIFAR-10 over SPR~\cite{kim2021continual}. We achieve same type of improvement for a more realistic scenario of asymmetric noise. These performance improvements can be attributed to our well-designed separation mechanism that can filter out noisy samples with high success rate. We show this in Table \ref{tab:exp_filtering}. Furthermore, keeping the noisy samples rather than discarding them creates the scope of SSL training. The SSL training with strong augmentations helps learning better representations/features than the conventional fully supervised fine-tuning. 

In addition, Table \ref{tab:exp_cifar100} shows the comparison of CNLL with other SOTA methods on CIFAR-100 dataset with random symmetric and super-class symmetric noise. As presented in Table \ref{tab:exp_cifar100}, CNLL outperforms the next best method by 17.2\% for 20\% random symmetric noise. This result is consistent for super-class symmetric noise. Figure \ref{fig:tsne} also resonates our claims as our method obtains a satisfactory separation of test samples from all classes. To obtain this clustering, we feed all test images to the trained DNN model only to get the output features. We get this output features from the CNN backbone of DNN. Then a technique named t-Distributed Stochastic Neighbour Embedding (t-SNE)~\cite{van2008visualizing} has been used to visualize these high dimensional features in a 2D map. The more separation and compactness of these clusters indicate the generalization performance of the DNN model. In both clean and noisy label cases, CNLL performs well in separating the test samples.  

\section{Scope of Future Work}\label{sec:future_work}
While we only considered noisy label in online task-free continual learning settings, there are other forms of continual learning that still needs to be explored. Furthermore, CNLL deals with symmetric and asymmetric cases, future studies can focus on the instance/part dependent label noise. To address this, SPR~\cite{kim2021continual} experiements with WebVision benchmark dataset~\cite{li2017webvision}. However, there are other real-world noisy label datasets such as CLothing1M~\cite{xiao2015learning} that contains a higher percentage of noisy labels. In future, we will focus on developing sophisticated algorithm for such noise scenarios. 
\section{Conclusion}
\label{sec:conclusion}
We propose a novel training scheme that consists of two phases: task-free sample separation and fine-tuning. By dynamically adjusting a class-specific binary mask, we apply a distance metric to separate the incoming data-stream into clean and noisy sets or buffers. The separation we proposed are dynamic and hyper-parameter independent. Next, we use semi-supervised fine-tuning instead of traditional fully supervised training. Due to the better design of noise cleansing mechanism and superior training scheme, CNLL merits over SOTA in 3 benchmark datasets. Through extensive experimentation, we show the effectiveness of our method as it achieves a 17.2\% accuracy improvement, for CIFAR100 with 20 tasks and 20\% noise, over the next best method. \newline         
\textbf{Acknowledgement:} This work was partly supported by the National Science Foundation under
Grant No. CCF-1718195.

{\small
\bibliographystyle{ieee_fullname}
\bibliography{egbib}
}

\end{document}